\documentclass[conference]{IEEEtran}
\IEEEoverridecommandlockouts
\usepackage{cite}
\usepackage{amsmath,amssymb,amsfonts}
\usepackage{algorithmic}
\usepackage{graphicx}
\usepackage{textcomp}
\usepackage{xcolor}
\usepackage{multirow}
\usepackage{booktabs}
\def\BibTeX{{\rm B\kern-.05em{\sc i\kern-.025em b}\kern-.08em
    T\kern-.1667em\lower.7ex\hbox{E}\kern-.125emX}}
\DeclareRobustCommand*{\IEEEauthorrefmark}[1]{%
	\raisebox{0pt}[0pt][0pt]{\textsuperscript{\footnotesize\ensuremath{#1}}}}

\begin{document}

\title{ReF-LLE: Personalized Low-Light Enhancement via Reference-Guided Deep Reinforcement Learning}

\author{
	\IEEEauthorblockN{
		Ming Zhao\IEEEauthorrefmark{1},
		Pingping Liu\IEEEauthorrefmark{2*}\thanks{* Corresponding author},
		Tongshun Zhang\IEEEauthorrefmark{2},
		Zhe Zhang\IEEEauthorrefmark{2}
	}
	\IEEEauthorblockA{
		\IEEEauthorrefmark{1}\textit{College of Software}, \textit{Jilin University, Changchun, 130012, China}
	}
	\IEEEauthorblockA{
		\IEEEauthorrefmark{2}\textit{College of Computer Science and Technology}, \textit{Jilin University, Changchun, 130012, China}
	}
	\IEEEauthorblockA{
		mingzhao23@mails.jlu.edu.cn, liupp@jlu.edu.cn, tszhang23@mails.jlu.edu.cn, zhezhang@mails.jlu.edu.cn
	}
}

\maketitle

\begin{abstract}
Low-light image enhancement presents two primary challenges: 1) Significant variations in low-light images across different conditions, and 2) Enhancement levels influenced by subjective preferences and user intent. To address these issues, we propose ReF-LLE, a novel personalized low-light image enhancement method that operates in the Fourier frequency domain and incorporates deep reinforcement learning. ReF-LLE is the first to integrate deep reinforcement learning into this domain. During training, a zero-reference image evaluation strategy is introduced to score enhanced images, providing reward signals that guide the model to handle varying degrees of low-light conditions effectively. In the inference phase, ReF-LLE employs a personalized adaptive iterative strategy, guided by the zero-frequency component in the Fourier domain, which represents the overall illumination level. This strategy enables the model to adaptively adjust low-light images to align with the illumination distribution of a user-provided reference image, ensuring personalized enhancement results. Extensive experiments on benchmark datasets demonstrate that ReF-LLE outperforms state-of-the-art methods, achieving superior perceptual quality and adaptability in personalized low-light image enhancement.
\end{abstract}


\section{Introduction}
\label{sec:intro}

The widespread use of mobile phones has lowered the barriers to photography. However, image quality in low-light conditions remains a significant challenge for computer vision tasks. This has led to extensive research into low-light image enhancement (LLIE) methods \cite{guo2016lime,zhang2019dual,ma2022toward,yang2023implicit, wang2022low}. 
Among these methods, Retinex-based methods \cite{fu2023learning,cai2023retinexformer,wu2022uretinex} approach low-light image enhancement by modeling it as $I = R \circ L$, aiming to recover the illumination $L$ to enhance the image. Other techniques \cite{huang2022deep,li2023embedding,wang2023fourllie} leverage Fourier frequency information, noting that luminance in Fourier space is primarily in the amplitude component. They enhance image luminance by amplifying this component and integrate this approach into neural networks, achieving impressive results.
\begin{figure}[t]
	\centering
	\includegraphics[width=1\columnwidth]{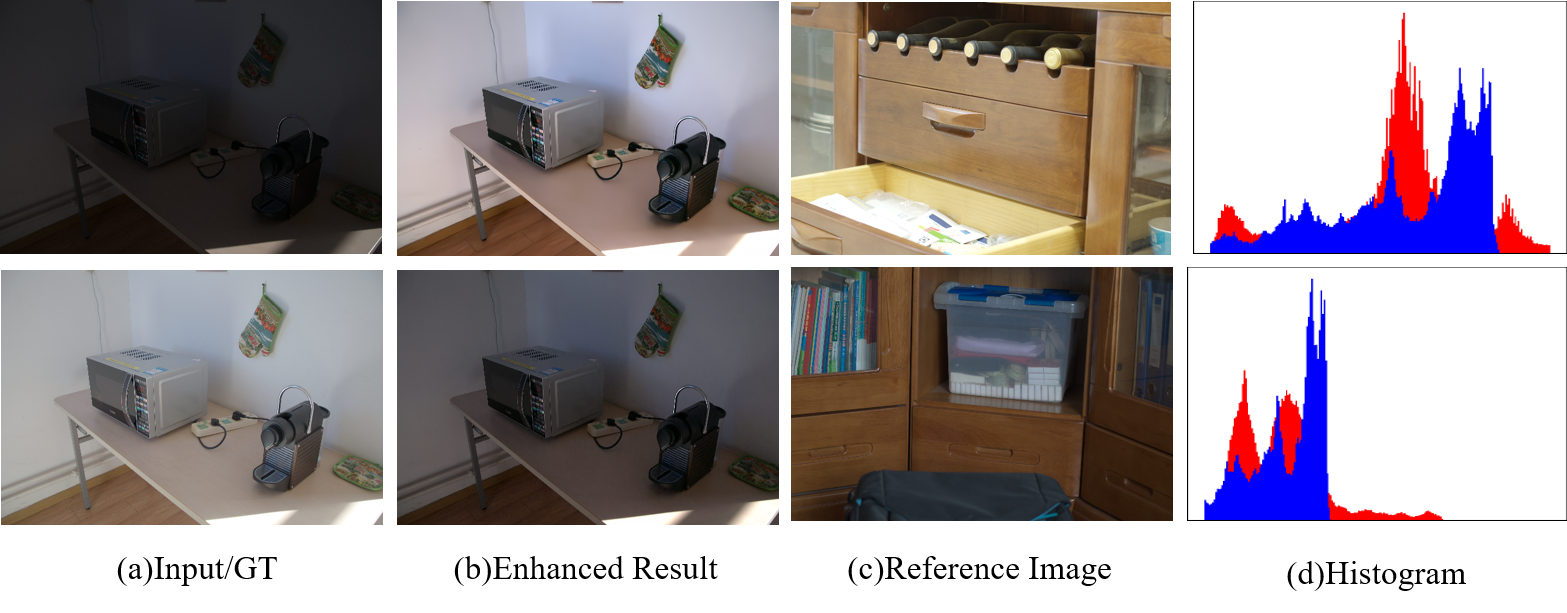} 
	\vspace{-0.6cm}
	\caption{The personalized results are guided by different reference images. (a) The input low-light image is placed above the ground truth; (b) the personalized enhancement results guided by the reference images presented in (c); (d) the brightness histograms comparing the reference images (red) and the enhanced results (blue).}
	\vspace{-0.6cm}
	\label{fig1}
\end{figure}

However, most existing methods adopt an end-to-end, point-to-point image enhancement approach, fitting the lighting distribution under a single dataset. Moreover, the LLIE task itself is complex and inherently subjective, as different users have distinct preferences for brightness fitting within their desired range. Therefore, fitting to complex lighting conditions and generating brightness variations that align with user preferences have become the focal points of this paper. 

Although several reinforcement learning-based methods (e.g., \cite{zhang2021rellie,li2023all}) have attempted to personalize image outputs by offering users multiple options, this approach can be cumbersome. Specifically, these methods generate multiple outputs by varying the number of iterative steps, which can increase the user's selection burden. As illustrated in Figure \ref{fig1}, LLIE methods should be more personalized, enabling adaptive enhancement of low-light images to align with the user's preferred light level.

Leveraging the strengths of addressing LLIE tasks in the Fourier frequency domain \cite{huang2022deep,li2023embedding,wang2023fourllie,tongshunzhang2024dmfourllie} while addressing the aforementioned limitations, we propose a novel method named ReF-LLE. This approach integrates deep reinforcement learning with the Fourier domain to tackle the challenges posed by complex lighting conditions. Additionally, during the inference phase, ReF-LLE employs a zero-frequency component prior as guidance, enabling an adaptive and personalized iterative process for low-light enhancement.

\begin{figure}[t]
	\centering
	\includegraphics[width=1\columnwidth]{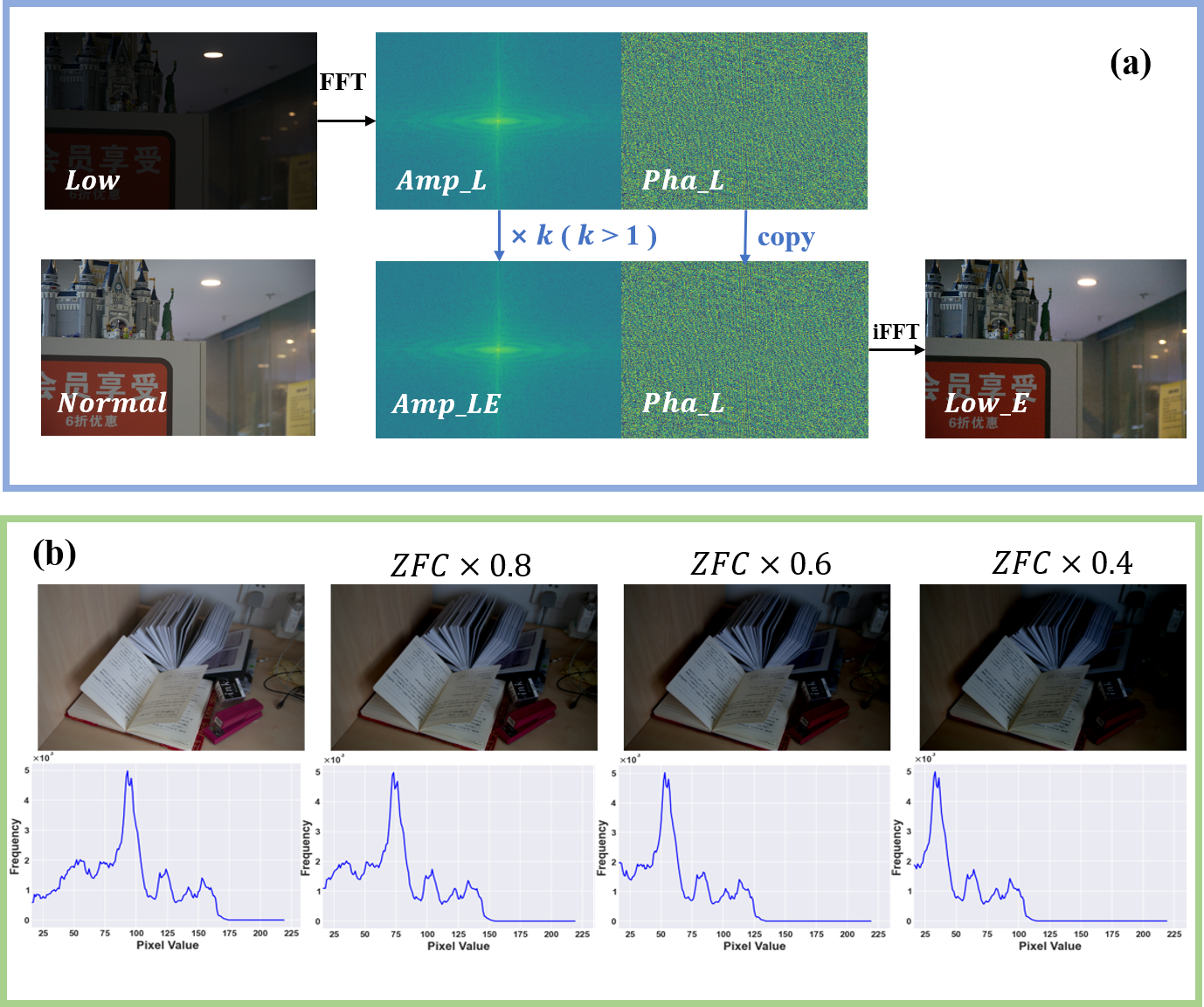} 
	\vspace{-0.6cm}
	\caption{Our motivations: (a) Amplifying the amplitude of a low-light image by a constant factor $k$ while preserving its phase information results in Low\_E, indicating that amplitude amplification can enhance low-light images; (b) The visual effects (top row) and Y-channel histograms (in the YCbCr color space, bottom row) demonstrate that varying reductions in ZFC lead to decreased overall image brightness, thereby highlighting a positive correlation with image brightness.}
	\vspace{-0.6cm}
	\label{fig2}
\end{figure}

In the Fourier space, \textbf{LLIE can be achieved by enhancing the amplitude component}(see Fig. \ref{fig2}(a)). Additionally, we scaled the \textbf{Z}ero-\textbf{F}requency \textbf{C}omponent (i.e., the centroid of the image after Fourier transform centering, referred to as ZFC in this paper;) of a normal-light image to varying degrees. Our observations indicate that \textbf{this component effectively represents the overall brightness of the image and shows a positive correlation with image brightness} (refer to Fig. \ref{fig2}(b)).

Based on the above observations, we integrate a reinforcement learning framework with Fourier theory to achieve unsupervised personalized low-light enhancement. ReF-LLE is engineered to process low-light or medium-light images as inputs, generating amplitude scaling coefficients at each step through a learned strategy. This learning process is directed by a series of meticulously crafted zero-reference image quality assessment rewards. In the inference phase, ReF-LLE employs a personalized adaptive iterative strategy, guided by the zero-frequency component, which enables the model to adaptively adjust low-light images to align with the illumination distribution of a user-provided reference image, ensuring personalized enhancement results. 
Our contributions are summarized as follows:
\begin{itemize}
	\item Recognizing the gap between realistic scenarios and the limitations of existing LLIE methods, we propose ReF-LLE, a deep reinforcement learning framework based on Fourier theory, aimed at achieving a more personalized LLIE approach.
	\item We conduct an in-depth exploration of the Fourier amplitude component, where the introduction of the zero-frequency component enables adaptive personalized low-light image enhancement.
	\item Our method has been thoroughly evaluated against state-of-the-art competitors using comprehensive experiments that focus on visual quality, no-reference, and full-reference image quality assessments. The results consistently demonstrate the clear superiority of ReF-LLE.
\end{itemize}

\begin{figure*}[t]
	\centering
	\includegraphics[width=2\columnwidth]{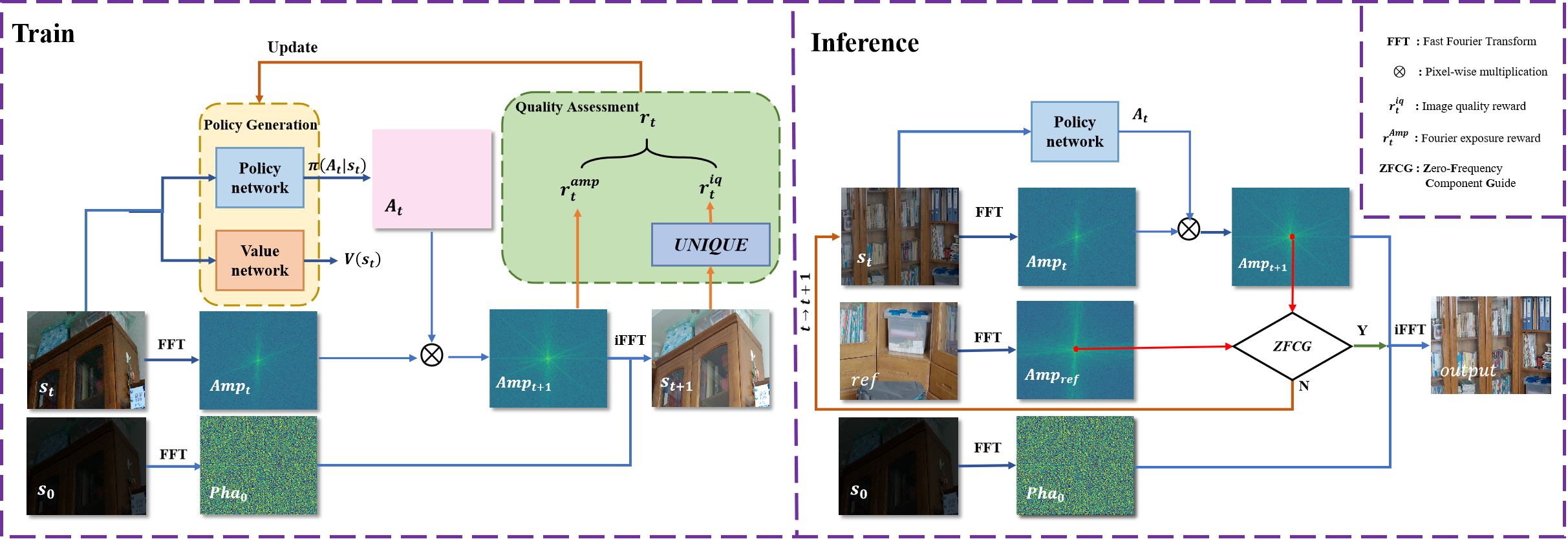} 
	\vspace{-0.2cm}
	\caption{Overall architecture of the proposed ReF-LLE.}
	\vspace{-0.6cm}
	\label{fig3}
\end{figure*}

\section{Problem Definition}
In Fourier space, an image is represented by amplitude and phase components. The amplitude mainly reflects intensity and lighting. Thus,  Low-Light Image Enhancement (LLIE) can be viewed as the optimization of the amplitude component of an image. We define LLIE as iterative linear enhancement of this component. In each iteration, we apply a linear enhancement to the magnitude, as expressed in the following equation:

\begin{equation}
	Amp_{t}(x)=A_{t}(x)Amp_{t-1}(x)
\end{equation}
where $Amp$ denotes the amplitude component of the image x. At each iteration, the optimal parameter mapping $A_t(x)$ is determined on a pixel-by-pixel basis for the amplitude information. The enhancement results from each iteration are then reconstructed into the RGB color space. This process can be expressed as:
\begin{equation}
	\begin{gathered}
		R_{t} = Amp_{t}(x) \times \cos\left(Pha_{0}(x)\right) \\
		I_{t} = Amp_{t}(x) \times \sin\left(Pha_{0}(x)\right) \\
		X_{t}(u,v) = R_{t} + jI_{t}\\
		x_{t} = \mathcal{F}^{-1}(X_{t}(u,v))
	\end{gathered}
\end{equation}
where $Pha_{0}(x)$ represents the phase component of the input low-light image, while our goal is to optimize its amplitude component during each iteration, keeping the phase component constant. The optimized real part $R_{t}$ and imaginary part $I_{t}$ are then reorganized into complex number $X_{t}(u,v)$ , which is then reconstructed into the RGB image $x_{t}$ using the inverse Fourier transform $\mathcal{F}^{-1}$.

\section{Methodology}
Our framework is illustrated in Fig. \ref{fig3}. In the training stage, we decompose low-light images using Fourier transform into amplitude and phase components. By optimizing the amplitude component for illumination and recombining it with the initial phase via inverse Fourier, we create enhanced images. A zero-reference evaluation provides feedback for learning. During inference, our framework uses a personalized adaptive iterative strategy that focuses on the Zero-Frequency Component (ZFC) to adjust the image's lighting levels. This iterative process ensures the lighting is adjusted to precisely match the user's desired brightness, accommodating individual preferences across various scenarios. 

\subsection{Policy Generation}
We utilize the deep reinforcement learning framework A3C\cite{mnih2016asynchronous}, which allows for the exploration of a variety of strategies through asynchronous threads. This methodology significantly enhances exploration efficiency and reduces the correlation between parameter updates, leading to more robust learning. The A3C method comprises two primary sub-networks: the policy network and the value network. At each step $t$ , the value network estimates the long-term discounted reward $R_t$ by evaluating the value $V(s_t)$ of a given state $s_t$(input image at the $ t $-th iteration). This estimation helps the agent to make informed decisions by balancing immediate rewards with future potential gains. The policy network and the value network work together to ensure that the agent considers both current and future returns in its decision-making process, thereby finding the optimal long-term strategy. Further details on the policy generation and optimization can be found in the supplementary material.

\subsection{Action Space Setting}
As previously discussed, we employ pixel-level linear enhancement for amplitude information. Specifically, the network decides on the appropriate action to modulate the amplitude based on the state $s_t$. The selection of the action range is crucial: too narrow a range limits the enhancement effect, while too wide a range increases the search space, thus raising computational complexity and training difficulty. To achieve a balance between efficiency and effectiveness, we define the action $A$ as follows: $A = e^\alpha$. This formulation allows for flexible adjustment of amplitude gain to meet the requirements of various application scenarios. Based on empirical evidence and experimental results, we set the parameter $\alpha$ to range between $[-0.1, 0.2]$ with a step size of $0.01$ for discretization. This configuration provides sufficient flexibility in enhancement while maintaining computational efficiency.
\subsection{Quality Assessment}
In deep reinforcement learning, the reward function is crucial for guiding model optimization through immediate feedback, aiming to maximize long-term rewards. In this work, we streamline the training process by using only two reward functions: Image quality reward and Amplitude exposure reward.
\subsubsection{Image quality reward.}
UNIQUE \cite{zhang2021uncertainty} is a recently proposed model for non-reference image quality evaluation that more closely aligns with subjective human perception, offering a more rational and insightful assessment approach. In this work, we employ UNIQUE to evaluate image quality in the Low-Light Image Enhancement task, using the input state $s_t$. UNIQUE assigns a quality score to the image, denoted as $\mathcal{S}(s_t)$. This score is then used as a reward in deep reinforcement learning, as described below:
\begin{equation}
	r_{t}^{iq}=\mathcal{S}_{}(s_{t}^{})-\mathcal{S}_{}(s_0)
\end{equation}
The image quality reward $r_t^{iq}$ is determined by measuring the UNIQUE score difference between the current state $s_t$ and the initial state $s_0$. A larger difference indicates a greater improvement in the optimized image quality.
\begin{table*}[htbp]
	\centering
	\caption{Quantitative comparison on the LOL, LOL-v2-Real, and LSRW-Huawei datasets the best results are highlighted in red, while the second-best results are highlighted in blue.}
	\vspace{-0.3cm}
	\resizebox{\textwidth}{!}{
		\begin{tabular}{|c|p{3.5em}|ccc|cccccc|cc|c|}
			\hline
			\multicolumn{2}{|c|}{\centering Methods} & \multicolumn{3}{c|}{\centering T} & \multicolumn{6}{c|}{\centering UL} & \multicolumn{3}{c|}{\centering DRL} \\
			\hline
			\multicolumn{1}{|c|}{\multirow{2}[2]{*}{\centering Datasets}} & \multirow{2}[2]{*}{\centering Metrics} & \multicolumn{1}{c}{LIME\cite{guo2016lime}} & \multicolumn{1}{c}{DUAL\cite{zhang2019dual}} & \multicolumn{1}{c|}{SDD\cite{hao2020low}} & \multicolumn{1}{c}{Zero-DCE\cite{guo2020zero}} & \multicolumn{1}{c}{SCI\cite{ma2022toward}} & \multicolumn{1}{c}{SCL-LLE\cite{liang2022semantically}} & \multicolumn{1}{c}{CLIP-LIT\cite{liang2023iterative}} & \multicolumn{1}{c}{FourierDiff\cite{lv2024fourier}} & \multicolumn{1}{c|}{RSFNet\cite{saini2024specularity}} & \multicolumn{1}{c}{ReLLIE\cite{zhang2021rellie}} & \multicolumn{1}{c|}{ALL-E\cite{li2023all}} & \multicolumn{1}{c|}{ReF-LLE} \\
			& \multicolumn{1}{c|}{} & \multicolumn{1}{c}{\textcolor{orange}{(TIP,16)}} & \multicolumn{1}{c}{\textcolor{orange}{(CGF,19)}} & \multicolumn{1}{c|}{\textcolor{orange}{(TMM,20)}} & \multicolumn{1}{c}{\textcolor{orange}{(CVPR,20)}} & \multicolumn{1}{c}{\textcolor{orange}{(CVPR,22)}} & \multicolumn{1}{c}{\textcolor{orange}{(AAAI,22)}} & \multicolumn{1}{c}{\textcolor{orange}{(ICCV,23)}} & \multicolumn{1}{c}{\textcolor{orange}{(CVPR,24)}} & \multicolumn{1}{c|}{\textcolor{orange}{(CVPR,24)}} & \multicolumn{1}{c}{\textcolor{orange}{(MM,21)}} & \multicolumn{1}{c|}{\textcolor{orange}{(IJCAI,23)}} & \multicolumn{1}{c|}{\textcolor{orange}{(Ours)}} \\
			\hline
			\multicolumn{1}{|c|}{\multirow{5}[2]{*}{LOL}}
			& PSNR$\uparrow$ & 14.22 & 14.02 & 13.34 & 14.86 & 14.79 & 12.42 & 12.39 & 17.56 & 19.39 & \textcolor{blue}{19.45} & 18.31 & \textcolor{red}{19.88} \\
			& PSNR$_y$$\uparrow$ & 16.20 & 15.97 & 15.14 & 16.76 & 16.48 & 14.58 & 14.14 & 19.66 & \textcolor{red}{22.17} & 21.73 & 20.37 & \textcolor{blue}{22.09} \\
			& SSIM$\uparrow$ & 0.521 & 0.519 & 0.634 & 0.559 & 0.522 & 0.520 & 0.493 & 0.607 & 0.755 & \textcolor{blue}{0.757} & 0.734 & \textcolor{red}{0.761} \\
			& LPIPS$\downarrow$ & 0.344 & 0.346 & 0.278 & 0.335 & 0.339 & 0.336 & 0.382 & 0.287 & 0.265 & \textcolor{blue}{0.210} & 0.231 & \textcolor{red}{0.191} \\
			\hline
			\multicolumn{1}{|c|}{\multirow{5}[2]{*}{LOL-V2-Real}}
			& PSNR$\uparrow$ & 17.14 & 16.95 & 16.64 & 18.06 & 17.30 & 15.40 & 15.18 & 16.86 & \textcolor{blue}{19.27} & 17.42 & 15.84 & \textcolor{red}{20.32} \\
			& PSNR$_y$$\uparrow$ & 19.31 & 19.10 & 18.47 & 20.34 & 19.41 & 17.18 & 17.05 & 19.02 & \textcolor{blue}{21.46} & 19.66 & 17.92 & \textcolor{red}{23.03} \\
			& SSIM$\uparrow$ & 0.537 & 0.535 & 0.678 & 0.574 & 0.534 & 0.556 & 0.529 & 0.603 & \textcolor{blue}{0.738} & 0.728 & 0.699 & \textcolor{red}{0.742} \\
			& LPIPS$\downarrow$ & 0.322 & 0.324 & 0.280 & 0.313 & 0.308 & 0.300 & 0.369 & 0.294 & 0.280 & \textcolor{blue}{0.223} & 0.264 & \textcolor{red}{0.189} \\
			\hline
			\multicolumn{1}{|c|}{\multirow{5}[2]{*}{LSRW-Huawei}}
			& PSNR$\uparrow$ & 15.34 & 14.10 & 14.60 & 16.34 & 15.70 & 13.46 & 13.56 & 16.87 & - & \textcolor{blue}{18.11} & 17.08 & \textcolor{red}{18.47} \\
			& PSNR$_y$$\uparrow$ & 16.63 & 15.37 & 15.84 & 17.68 & 17.01 & 14.72 & 14.79 & 18.69 & - & \textcolor{blue}{19.93} & 18.61 & \textcolor{red}{20.24} \\
			& SSIM$\uparrow$ & 0.348 & 0.435 & 0.517 & 0.475 & 0.436 & 0.427 & 0.427 & 0.511 & - & \textcolor{blue}{0.549} & \textcolor{red}{0.557} & 0.529 \\
			& LPIPS$\downarrow$ & 0.397 & 0.407 & 0.494 & 0.398 & 0.381 & 0.390 & 0.408 & \textcolor{red}{0.366} & - & 0.429 & 0.430 & \textcolor{blue}{0.381} \\
			\hline
		\end{tabular}
	}
	\label{tab1}
\end{table*}

\begin{figure*}[t]
	\centering
	\includegraphics[width=2\columnwidth]{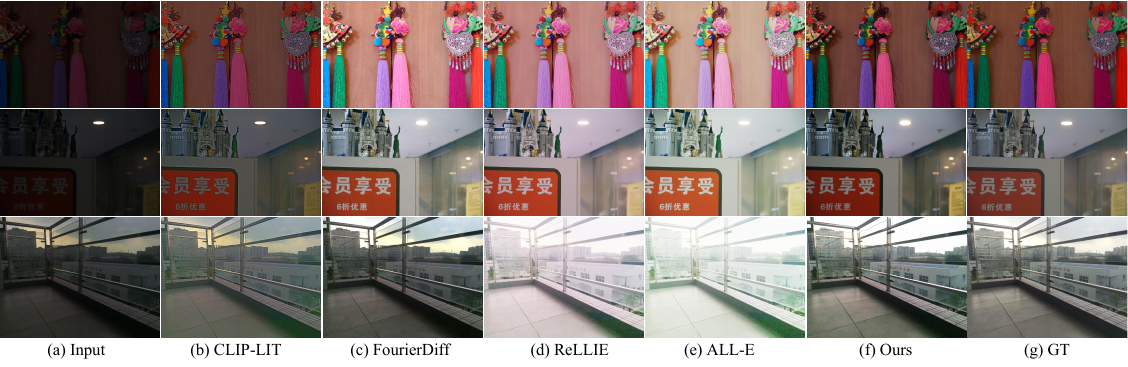} 
	\vspace{-0.3cm}
	\caption{Visual comparison with state-of-the-art low-light image enhancement methods on the LOL(top row), LOL-v2-Real(middle row), and LSRW-Huawei(bottom row) datasets.}
	\vspace{-0.6cm}
	\label{fig10}
\end{figure*}

\begin{figure}[t]
	\centering
	\includegraphics[width=1\columnwidth]{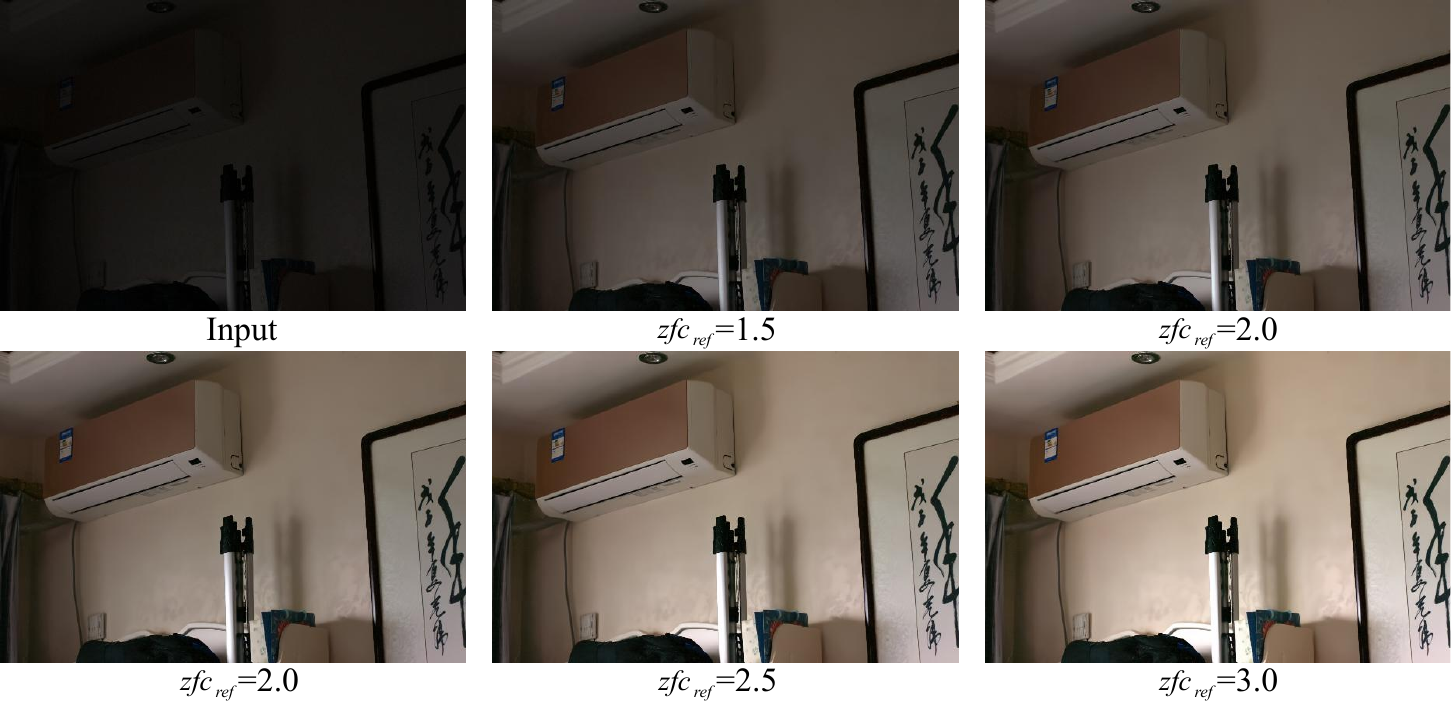} 
	\vspace{-0.6cm}
	\caption{Examples of customized low-light image enhancement by adjusting the global-illumination parameter $zfc_{ref}$.}
	\vspace{-0.2cm}
	\label{fig11}
\end{figure}
\subsubsection{Amplitude exposure reward.}
To accurately control the global exposure level, we devised the amplitude exposure reward: 
\begin{equation}
	r_{t}^{amp}=|\frac{\bar{zfc}}{zfc}_t-1|
\end{equation}
Here, ${zfc}_t$ denotes the zero-frequency component of state $s_t$ , and $\bar{zfc}$ is a hyperparameter that controls the desired zero-frequency component to be achieved within a fixed number of iterations during the training process. A detailed discussion of the zero-frequency component will be provided in the following subsection.

Hence, for a given enhanced image, the immediate reward $r_t$ 
at a current state $s_t$ is:
\begin{equation}
	r_{t}=w_{iq}r_{t}^{iq}-w_{amp}r_{t}^{amp}
\end{equation}
where $w_{iq}$ and $w_{amp}$ are tunable hyperparameters. The goal of reinforcement learning is to maximize the total discounted reward $R_t$
in Eq.(8) with this immediate reward $r_t$.

\subsection{Zero-frequency component Guide}
The Fourier transform converts information from the spatial domain into the frequency domain. As outlined in \cite{jahne2005digital}, zero-frequency component (ZFC) represents the sum or average light intensity of all pixel values in the image, thereby directly reflecting the overall brightness level. Since the zero-frequency component does not contain any frequency-varying information, it only captures the constant part of the image—namely, the global illumination and the underlying luminance distribution. Consequently, we adopt the ZFC as a prior for global illumination. Ref-LLE adaptively iterates to progressively align the ZFC of the enhanced image ($zfc_t$
) with the ZFC of the reference image (
$zfc_{ref}$) provided by the user. We use a formula similar to the one presented in $r^{amp}$ to measure the brightness difference between the two.
\begin{figure}[t]
	\centering
	\includegraphics[width=1\columnwidth]{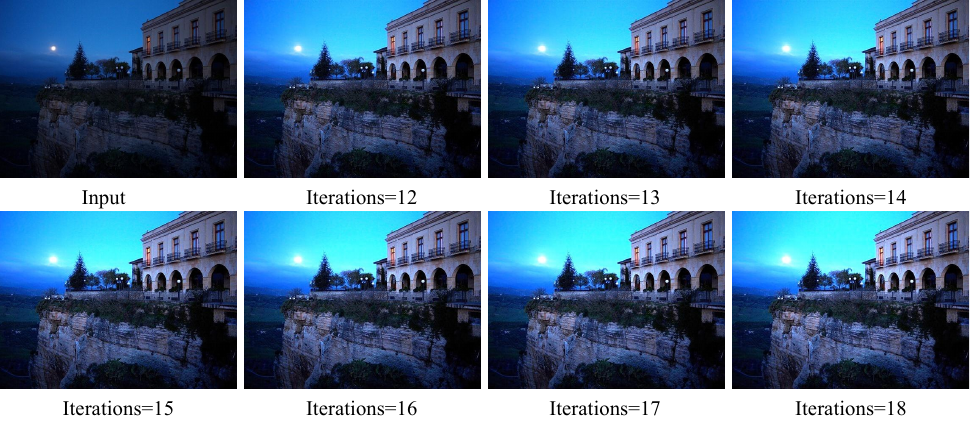} 
	\vspace{-0.1cm}
	\caption{Examples of customized low-light image enhancement by setting different number of iterations.}
	\vspace{-0.2cm}
	\label{fig12}
\end{figure}
\begin{figure*}[h]
	\centering
	\includegraphics[width=2\columnwidth]{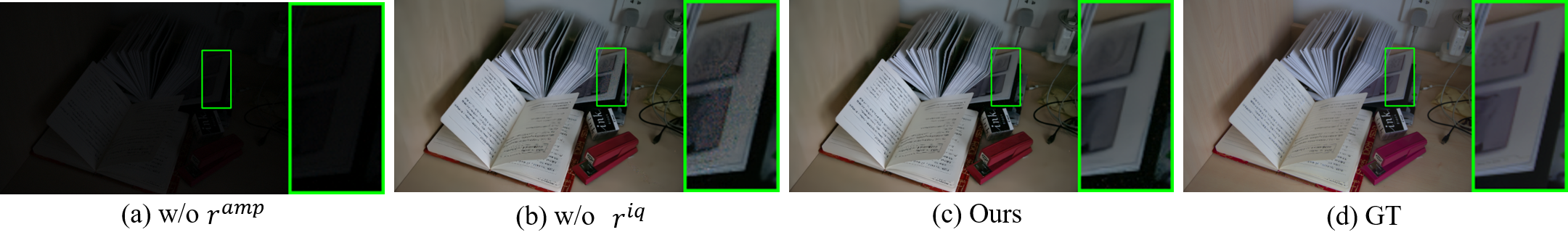} 
	\vspace{-0.3cm}
	\caption{Visualization of the contribution of each reward (Amplitude exposure reward $r^{amp}$ and Image quality reward $r^{iq}$); please zoom in for better visualization.}
	\label{fig9}
\end{figure*}

\section{Experiments and Results}
\subsection{Experiments Setting}
\subsubsection{Datasets and baselines.}
We trained our model using the LOL\cite{wei2018deep} dataset, which was captured in real scenes with varying exposure times and ISO values. The dataset includes 485 pairs of low/normal light images for training and 15 pairs for testing. We assessed our model on the LOL, LOL-v2-Real \cite{yang2021sparse}, and LSRW-Huawei \cite{hai2023r2rnet} datasets. Compared to the original LOL dataset, LOL-v2-Real is larger and more diverse, offering a more comprehensive evaluation of model performance. The LSRW-Huawei dataset, captured in real-world conditions using different devices, consists of 3,150 training image pairs and 20 test image pairs. Additionally, we also evaluated ReF-LLE on five unpaired datasets: DICM\cite{yang2020advancing}(64 images), LIME\cite{lee2013contrast}(10 images), MEF\cite{guo2016lime}(17 images), NPE\cite{wang2013naturalness}(85 images) and VV\cite{ma2015perceptual}(24 images).
In this paper, we compare our ReF-LLE method with nine other state-of-the-art unsupervised low-light image enhancement methods. These include traditional methods such as LIME\cite{guo2016lime}, DUAL\cite{zhang2019dual} and SDD\cite{hao2020low}, as well as six unsupervised methods: Zero-DCE\cite{guo2020zero}, SCI\cite{ma2022toward}, SCL-LLE\cite{liang2022semantically}, CLIP-LIT\cite{liang2023iterative},  RSFNet\cite{saini2024specularity} and FourierDiff\cite{lv2024fourier}. We also compare against reinforcement learning methods ReLLIE\cite{zhang2021rellie} and ALL-E\cite{li2023all}. It should be noted that RSFNet is not open source, so we used the results reported in their paper for comparison. The results for the other methods were obtained by reproducing their official codebases.
\subsubsection{Implementation details.}
We implemented the proposed method using the PyTorch framework. During the training process, we set the total number of steps to 10. In the testing phase, ReF-LLE performs adaptive iterations based on our pre-set parameters. We set the reward weights as $w_{iq} = 1000$ and $w_{amp} = 60$. For the reward $r^{amp}$, we set the hyperparameter $\bar{zfc}$ to $2.5 \times 10^5$. The maximum number of rounds during training is set to 10,000, and the batch size is 2. The discount factor $\gamma$ is 0.95, and the learning rate is set to 0.002. All experiments were performed on a NVIDIA RTX A5000 GPU.
\begin{table}[]
	\vspace{-0.3cm}
	\caption{NIQE scores on LIME, VV, DICM, NPE, and MEF
		datasets.The top results are highlighted in red and the second-best in blue. "AVG" denotes the average NIQE scores across these five datasets.}
	\begin{tabular}{ccccccc}
		\hline
		Methods     & LIME & VV   & DICM & NPE  & MEF  & AVG   \\ \hline
		Zero-DCE    & 3.77 & 3.21 & 3.56 & 3.93 & 3.28 & 3.55  \\
		SCI         & 4.21 & 2.92 & 4.11 & 4.46 & 3.63 & 3.87  \\
		SCL-LLE     & 3.78 & 3.16 & 3.51 & 3.88 & 3.31 & 3.53  \\
		FourierDiff & \textcolor{blue}{3.29} & 3.19 & 3.26 & \textcolor{blue}{3.06} & \textcolor{blue}{3.03} & 3.17  \\
		RSFNET      & 3.8  & \textcolor{red}{1.96} & \textcolor{red}{3.23} & 3.31 & \textcolor{red}{3.00} & \textcolor{blue}{3.06}  \\
		ALL-E       & 3.78 & 3.08 & 3.49 & 3.85 & 3.32 & 3.504 \\ \hline
		Ours        & \textcolor{red}{3.18} & \textcolor{blue}{2.67} & \textcolor{blue}{3.25} & \textcolor{red}{2.99} & \textcolor{red}{3.00} & \textcolor{red}{3.02}  \\ \hline
	\end{tabular}
	\label{tab3}
\end{table}

\subsubsection{Metrics.}
As full-reference metrics (requiring real reference values), we utilize the Peak Signal-to-Noise Ratio (PSNR), the Structural Similarity Index Metric (SSIM)\cite{wang2004image}, and the Learned Perceptual Image Patch Similarity (LPIPS)\cite{zhang2018unreasonable}. PSNR is evaluated on both the single Y channel (from YCbCr) and the multi-channel (RGB) representations, respectively. For no-reference assessment (without ground truth), we report Naturalness Image Quality Evaluator (NIQE)\cite{mittal2012making}.

\subsection{Comparison}
We compare three model-driven traditional methods: LIME\cite{guo2016lime}, DUAL\cite{zhang2019dual}, and SDD\cite{hao2020low}; five state-of-the-art unsupervised methods: Zero-DCE\cite{guo2020zero}, SCI\cite{ma2022toward}, SCL-LLE\cite{liang2022semantically}, CLIP-LIT\cite{liang2023iterative}, and FourierDiff\cite{lv2024fourier}; as well as two reinforcement learning methods: ReLLIE\cite{zhang2021rellie} and ALL-E\cite{li2023all}. Qualitative comparisons are illustrated in Fig. \ref{fig10}. We conducted a qualitative comparison between the reinforcement learning approach and the state-of-the-art unsupervised methods from the past two years. Notably, CLIP-LIT exhibits underexposure, while ReLLIE and ALL-E show varying degrees of overexposure. Additionally, the enhancement results of FourierDiff are accompanied by the introduction of noise. However, our method achieves the closest approximation to the true values in terms of exposure levels, color fidelity, and overall performance. Quantitative comparisons, presented in Table \ref{tab1} (paired datasets) and Table \ref{tab3} (unpaired datasets) indicate that our method attains optimal or sub-optimal results across almost all metrics. 

\subsection{Visualization of Personalized LLIE}
Flexibility and scalability refer to the system's ability to efficiently adapt to varying user preferences with minimal manual intervention. The proposed approach leverages the Ref-LLE algorithm, which empowers users to personalize the image enhancement process according to their specific aesthetic requirements. This customization is facilitated through three key interactive options: (1) the provision of a preference reference image (as illustrated in Fig. \ref{fig1}), (2) the configuration of the global-illumination parameter (refer to Fig. \ref{fig11}), and (3) the specification of the number of iterations(refer to Fig. \ref{fig12}). These options allow users to finely tune the enhancement procedure, ensuring that the system can meet diverse aesthetic expectations while maintaining its robustness and versatility.

\begin{figure}[t]
	\centering
	\includegraphics[width=1\columnwidth]{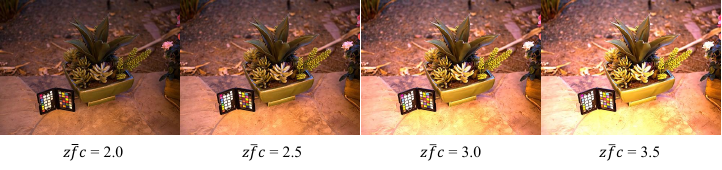} 
	\vspace{-0.5cm}
	\caption{The examples are generated with the same number of iterations under different settings of $\bar{zfc}$.}
	\vspace{-0.5cm}
	\label{fig13}
\end{figure}
\begin{table}[h]
	\centering
	\vspace{-0.3cm}
	\caption{Ablation studies of the contribution of reward functions.}
	\vspace{-0.3cm}
	\begin{minipage}{0.5\textwidth} 
		\centering
		\begin{tabular}{cc|ccc}
			\hline
			\multicolumn{2}{c|}{Settings} & \multicolumn{3}{c}{Metics}                                                       \\ \hline
			$r^{iq}$            & $r^{amp}$            & PSNR                      & SSIM                     & LPIPS                    \\ \hline
			\checkmark             &               & 7.31                      & 0.14                     & 0.651                     \\
			& \checkmark             & 19.60                     & 0.74                     & 0.206                     \\ \hline
			\multicolumn{2}{c|}{Setting1}  & 19.85                     & 0.74                     & 0.210                     \\
			\multicolumn{2}{c|}{Setting2}  & 19.84 & 0.74 & 0.209 \\
			\multicolumn{2}{c|}{Setting3}  & 19.82 & 0.68 & 0.219 \\ \hline
			\multicolumn{2}{c|}{Ours}     & 19.88                     & 0.76                     & 0.191                     \\ \hline
		\end{tabular}
		\label{tab:addlabel}%
	\end{minipage}
	\label{tab2}
\end{table}%
\vspace{-0.2cm}
\subsection{Ablation Study}
To investigate the effectiveness of the proposed reward functions (including $r^{amp}$ and $r^{iq}$)) and hyperparameter $\bar{zfc}$ in $r^{amp}$, we conducted an ablation study on the unsupervised LLIE and summarized the results in Table \ref{tab2}. Fig. \ref{fig9} provides a qualitative example illustrating the impact of each reward on the output. The results reveal that the absence of $r^{iq}$ leads to inadequate image enhancement, as UNIQUE still maintains a high rating in dark images. Conversely, the lack of $r^{amp}$ results in speckled and blurred images. For $\bar{zfc}$, we set the values to 2.0, 3.0, and 3.5 as Settings 1–3, respectively. The best performance was achieved when $\bar{zfc}$ was set to 2.5 (ours). As $\bar{zfc}$ controls the desired zero-frequency component to be achieved within a fixed number of iterations during the training process, setting a larger value of $\bar{zfc}$ causes ReF-LLE to prioritize actions that lead to a greater degree of enhancement at each iteration. Fig. \ref{fig13} visualizes the results obtained after the same number of iterations under different settings.

\section{Conclusion}
We introduce a novel personalized low-light enhancement method, ReF-LLE. By investigating the impact of Fourier amplitude information on image illumination, we employ deep reinforcement learning to iteratively optimize Fourier frequency data, allowing ReF-LLE to effectively process images with varying brightness levels. Building on this foundation, we present a comprehensible enhancement process guided by a strategy based on the zero-frequency component, resulting in customized outcomes that align with users' aesthetic preferences. Extensive qualitative and quantitative experiments confirm that ReF-LLE outperforms existing methods in unsupervised low-light image enhancement scenarios. 

\bibliographystyle{IEEEbib}
\bibliography{icme2025references}

\vspace{12pt}

\end{document}